\documentclass[runningheads]{llncs}
\usepackage{graphicx}
\usepackage{amsmath,amssymb} 
\usepackage{color}
\usepackage{multirow}
\usepackage{array}
\usepackage[width=122mm,left=12mm,paperwidth=146mm,height=193mm,top=12mm,paperheight=217mm]{geometry}
\begin{document}
\pagestyle{headings}
\mainmatter
\def\ECCV18SubNumber{****}  

\title{SphereReID: Deep Hypersphere Manifold Embedding for Person Re-Identification} 

\titlerunning{\quad}

\authorrunning{\quad}

\author{Xing Fan, Wei Jiang\thanks{Corresponding author, Email address: jiangwei\_zju@zju.edu.cn (Wei Jiang)},  Hao Luo, Mengjuan Fei}
\institute{\it{Institute of Cyber-Systems and Control, Zhejiang University,\\ Hangzhou 310027, China}\\
\{xfanplus, jiangwei\_zju, haoluocsc, feimengjuan\}@zju.edu.cn
}

\maketitle

\begin{abstract}
Many current successful Person Re-Identification(ReID) methods train a model with the softmax loss function to classify images of different persons and obtain the feature vectors at the same time. However, the underlying feature embedding space is ignored. In this paper, we use a modified softmax function, termed Sphere Softmax, to solve the classification problem and learn a hypersphere manifold embedding simultaneously. A balanced sampling strategy is also introduced. Finally, we propose a convolutional neural network called SphereReID adopting Sphere Softmax and training a single model end-to-end with a new warming-up learning rate schedule on four challenging datasets including Market-1501, DukeMTMC-reID, CHHK-03, and CUHK-SYSU. Experimental results demonstrate that this single model outperforms the state-of-the-art methods on all four datasets without fine-tuning or re-ranking. For example, it achieves 94.4\% rank-1 accuracy on Market-1501 and 83.9\% rank-1 accuracy on DukeMTMC-reID. The code and trained weights of our model will be released.

\end{abstract}

\section{Introduction}

Person re-identification is the task of identifying bounding box images of the same person from non-overlapping camera views. Given a probe image, we need to retrieve all images of the same person ID in gallery images.

Person re-identification has many practical applications such as video surveillance for public security, and thus attracts much research attention in the computer vision community. With the utilization of deep convolution neural networks (CNNs) \cite{AlexNet_2012} in recent years, ReID performance has made significant progress. However, some problems remain to be solved owing to the challenges of ReID, including changes in camera viewpoints, illumination changes, human pose variation and occlusion.

\begin{figure}
\centering
\includegraphics[width=.8\linewidth]{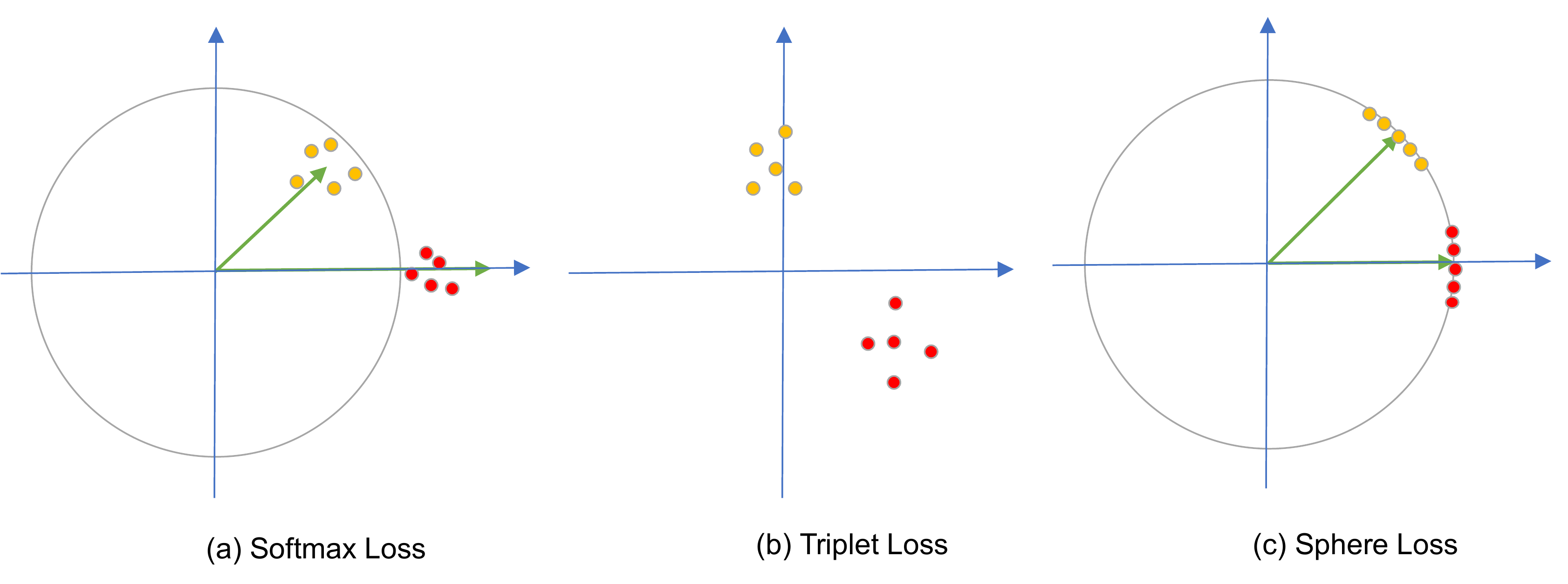}
\caption{Two-dimensional visualization of sample distribution in the embedding space supervised by (a)Softmax Loss, (b)Triplet Loss \cite{TripletLoss}, and (c)Sphere Loss. Yellow and red points represent embedding features from two different classes. }
\label{fig:visualizaion}
\end{figure}

Most of the current ReID approaches can be categorized into two types: feature-based or metric-based. Extracting features from input images and seeking a metric for comparing these features across images are the two main components of person re-identification. Some hand crafted features such as scale invariant feature transforms(SIFT) features \cite{SIFT_ICCV_2013,SIFT_CVPR_2013} and local maximal occurrence(LOMO) features \cite{LOMO_2015} have been used to represent the a person's appearance.

With the success of deep learning, CNN-based methods have been proposed for ReID to automatically learn the feature representations from the training data. These methods \cite{STN_2017,Spindle_2017,LearnedPart_2017,GLAD_2017} often model ReID as a classification problem and consider images from a specific person ID as a class. Then the softmax cross-entropy loss is applied to supervise the training procedure. Simultaneously, as a by-product, feature vectors before the last fully connected layer are extracted as the final image features. It corresponds with intuition that when a feature vector can be used to classify a person ID correctly, it is a good representation of that person's appearance. However, without explicit constraints on the feature space distribution, the learned feature mapping may not be optimal.
As shown in Fig.~\ref{fig:visualizaion}(a), there is no constraint on the distribution in the embedding space, which leads to a general spread.

To overcome the aforementioned drawbacks of feature-based works, metric-based methods \cite{Gate_2016,LSTM_2016,Triplet_ReID_2015,TripletHard,Quadruplet} have been proposed to learn an embedding of the original images that satisfies some specified conditions.  For example, triplet loss \cite{TripletLoss} requires the distance of samples from the same class to be less than that of samples from different classes by a pre-defined threshold, which pulls the instances of the same person closer and simultaneously pushes the instances belonging to different persons away from each other in the embedding space. Then the learned model is used for feature mapping of test images, and the extracted features can be compared using the Euclidean distance criterion. However, the range of each dimension is from minus infinity to plus infinity, and the feature of each dimension only lies within a small interval, as shown in Fig.~\ref{fig:visualizaion}(b).
Consequently, the target embedding space may not be fully utilized.

In this paper, we propose a novel metric-based person re-identification network called SphereReID, which adopts a new function called Sphere Loss to supervise the training process. Softmax cross-entropy is the basic loss function for the classification task. Despite the widespread use of softmax, whether it is the optimal loss function for classification is still uncertain. With the re-examination of softmax in the face recognition community \cite{L_Softmax_2016,A_Softmax_2017,AAM_Softmax_2018,CosFace_2018,ArcFace_2018}, some valuable insights have been obtained. Motivated by their works, we adopt a modified softmax loss function called Sphere Loss, which classifies image samples from different persons and restrains the distribution of sample embeddings on a hypersphere manifold at the same time.

To the best of our knowledge, this is the first time a person image has been mapped onto a hypersphere manifold for person re-identification. To this end, feature normalization and weight normalization are introduced. After elimination of different norms, the classification will only rely on the angle between the embedding vector and the target class weight vector, which has a more clear geometric interpretation in the embedding space, as shown in Fig.~\ref{fig:visualizaion}(c), where embedding features lie on a hypersphere manifold. Compared with Euclidean space embedding, SphereReID maps images on the surface of a hypersphere, which limits the possible space distribution to a restricted angular space. Thus, the target embedding space can be fully exploited and we can train a network to classify images from different persons and simultaneously regulate the target embedding distribution. Furthermore, the implementation of Sphere Loss is simple and the code will be released.

One issue with person re-identification is that there are many datasets \cite{VIPeR_2008,PRID_2011,iLIDS_VID_2016,CUHK_03,Market_1501_2015,CUHK_SYSU,Mars_2016,DukeMTMC_reID_2017,PRW} and everay labelled person has an indefinite quantity of images, thus sampling amount bias always exists. Further, some ReID datasets are image-based \cite{VIPeR_2008,CUHK_03,Market_1501_2015,CUHK_SYSU,DukeMTMC_reID_2017,PRW} whereas some are video-based \cite{PRID_2011,iLIDS_VID_2016,Mars_2016} consisting of a lot of consecutive images frames, which makes the images per person ID even more diverse. A softmax supervised classification approach suffers from the sample amount bias and end up with an inferior performance. In this paper, a balanced sampling strategy is introduced in the training process, and every mini-batch is generated by sampling a specific number of each person ID with a specific number of images which alleviates the sample amount bias problem.

With a new warming-up learning rate schedule, we train a single SphereReID model end-to-end without fine-tuning on four ReID datasets, and this single model outperforms the state-of-the-art methods on all the four datasets and achieves rank-1 accuracy 94.4\% on Market-1501 \cite{Market_1501_2015}, 83.9\% on DukeMTMC-reID \cite{DukeMTMC_reID_2017}, 93.1\% on CUHK03 \cite{CUHK_03} and 95.4\% on CUHK-SYSU \cite{CUHK_SYSU}.

The contribution of our work is three-fold:

- First, we introduce a new classification loss function called Sphere Loss modified from the softmax loss function, which can supervise the model to classify images of different persons and learn an embedding on a hypersphere manifold simultaneously.

- Second, a balanced sampling strategy is adopted to eliminate the sample amount bias and further facilitate the model performance without additional computational overhead. During training, a warming-up learning rate schedule also be used to bootstrap the network, which leads to a better convergence point.

- Finally, we propose a novel network called SphereReID adopting Sphere Loss. Extensive experiments on four datasets demonstrate the effectiveness of our proposed model.

\section{Related Works}

\subsubsection{Feature-Based ReID.}

Some hand crafted features such as scale invariant feature transforms(SIFT) features \cite{SIFT_ICCV_2013,SIFT_CVPR_2013}, Local Binary Patterns(LBP) features \cite{LBP_2014}, and local maximal occurrence(LOMO) features \cite{LOMO_2015} have been used to represent a person's appearance. With the rise of deep learning, automatically learning feature representations have been used for the ReID task and significant progress has been made as a result. Features extracted by a pre-trained CNN on a large annotated dataset, \emph{e.g.}, ImageNet, have been proven to be strong off-the-shelf descriptors for various recognition tasks, and Matsukawa \textit{et al.} \cite{CLS_ATTR_2016} present CNN features for person re-identification fine-tuned on a pedestrian attribute dataset. To extract fine-grained part feature, Varior \textit{et al.} \cite{Gate_2016} present a gate structure, while LSTM \cite{LSTM_1997} is introduced in \cite{LSTM_2016,LSTM_Attention_2017,DeepPerson_2017} to learn horizontal local features. Additonally, Sun \textit{et al.} \cite{Refined_2017} use horizontal stripes and Li \textit{et al.} \cite{STN_2017} use a Spatial Transform Networks (STN) \cite{STN} subnet to localize the refined body parts and Zhao \textit{et al.} \cite{LearnedPart_2017} learn the parts automatically through a mask predictor. Furthermore, Yao \textit{et al.} \cite{PartLoss_2017} represent different body parts by directly clustering feature maps based on the location of their maximum responses. As the human body is highly structured with known key points, external skeleton models have also been used for predicting different body regions in \cite{GLAD_2017,PoseBox_2017,PoseDriven_2017,Spindle_2017}.

\subsubsection{Metric-Based ReID.}
Along with feature-based methods, there are some approaches to ReID that use metric learning, which formulate the person re-identification as a supervised distance metric learning problem. Traditional metric learning methods like the Keep It Simple and Straightforward Metric(KISSME) \cite{KISSME} and cross-view quadratic discriminant analysis (XQDA) \cite{LOMO_2015} learn a transformation matrix of features. Nowadays, however, the community pays more attention to the loss function of a network. Instead of the softmax classification loss function, contrastive loss \cite{ContrastiveLoss} is used to supervise a Siamese network in \cite{LSTM_2016,Gate_2016}. Motivated by FaceNet \cite{FaceNet}, a convolutional neural network used to learn an embedding for faces, triplet loss \cite{TripletLoss} is used in \cite{ECCV_TRI_2014,Triplet_ReID_2015} to optimize in the embedding space such that embedding features for the same identity are closer to each other than those of different identities. Cheng \textit{et al.} \cite{ImprovedTripletLoss} propose an improved triplet loss by introducing another
pull term into the loss, penalizing large distances between positive embeddings. Quadruplet loss proposed in \cite{Quadruplet}, adds another pull term for the distance between negative pairs, which can lead to a model with a larger inter-class variation and a smaller intra-class variation. Generation of samples of triplets or quadruplets will remain a challenge, as easy samples will lead to a degeneration and too difficult samples may result in gradient explosion. To solve this problem, Hermans \textit{et al.} \cite{TripletHard} propose a batch hard sampling strategy.

\subsubsection{Other ReID Methods.}
Xiao \textit{et al.} \cite{DomainGuidedDropout} propose a domain guided dropout algorithm to improve the feature
learning procedure for multiple ReID datasets. Geng \textit{et al.} \cite{ConsistentDropout} introduce pairwise-consistent dropout for the pairwise verification loss layers, that is, each pair of compared training data points share the same dropout mask. AlignedReID \cite{AlignedReID} introduces a feature matching method to align different body parts. And DarkRank \cite{DarkRank} shows that a powerful teacher model can significantly help the training of a smaller and faster student network for ReID. Re-ranking methods \cite{Re_Ranking_1,Re_Ranking_2} can also be used to rearrange the original ranking list to further improve the accuracy. Generative adversarial nets (GAN) \cite{GAN} have also been proven to be effective, and can also be exploited for the ReID task. PTGAN \cite{PersonTransferGAN} proposes a Person Transfer Generative Adversarial Network (PTGAN) to bridge the domain gap between different datasets which relieves the expensive costs of annotating new training samples. Pose-normalization GAN (PN-GAN) \cite{PoseNormalized} proposes a deep person image generation model for synthesizing realistic person images conditional on pose.

\subsubsection{Softmax Re-examination in Face Recognition.}
The face recognition community has re-examined the meaning of softmax \cite{L_Softmax_2016,A_Softmax_2017,AAM_Softmax_2018,CosFace_2018,ArcFace_2018} and obtained valuable insights. Large-margin softmax (L-Softmax) loss is introduced in \cite{L_Softmax_2016}, and it maps the cosine value between feature vectors and the weight vector to a monotonically decreasing function with a large margin. The angular softmax (A-Softmax) loss proposed by \cite{A_Softmax_2017} enables convolutional neural networks to learn angularly discriminative features and weight normalization is introduced. In \cite{AAM_Softmax_2018,CosFace_2018,ArcFace_2018}, feature normalization is also applied, which makes the classification results only depend on the angle between the feature vector and weight vector.

\section{Our Approach}

\subsection{Softmax Loss}

\begin{figure}
\centering
\includegraphics[width=.8\linewidth]{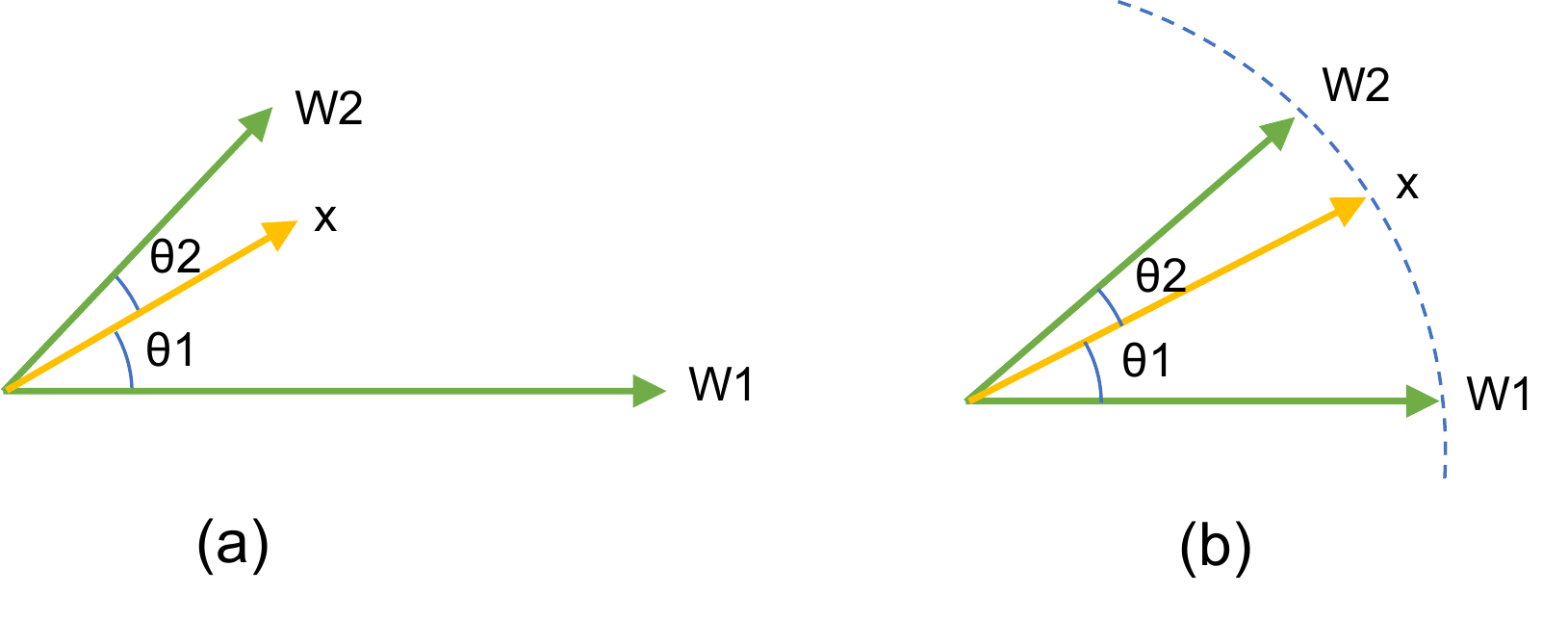}
\caption{Geometrical interpretation of (a)Softmax Loss and (b)Sphere Loss. Yellow arrows represent embedding feature vectors and green arrows represent class center weight vectors of two different classes.}
\label{fig:interpretation}
\end{figure}

In this section, we will discuss the meaning of the softmax loss function. Softmax is commonly used for classification task. Given an input feature vector $x_i$ with its corresponding
label $y_i$, it can be formulated as follows:

\begin{align}
  L_{softmax} = -\frac{1}{N}\sum_{i=1}^{N}\log\frac{e^{z_{y_i}}}{\sum_{j=1}^{C}e^{z_j}}
\label{eq:softmax}
\end{align}

where $N$ is the number of training samples and $C$ is the number of classes. $z_j$ is activation of the $j$-th neuron in a fully connected layer with weight vector $W_j$ and bias $b_j$. There are a total of $C$ neurons, and each neuron outputs the score $z_j$ of the corresponding $j$-th class. We fix the bias $b_j = 0$ for simplicity, and as a result we can formulate $z_j$ as follows:

\begin{align}
  z_j = W_j^Tx = \|W_j\| \|x\|\cos{\theta_j}
\end{align}

where $\theta_j$ is the angle between $W_j$ and $x$. As shown in Fig.~\ref{fig:interpretation}(a), for an embedding feature vector $x$, and learnable weights $W_1$ and $W_2$ which serve as the class center, both the feature vector and weight vector influence the output scores. For a binary classification, when $z_1>z_2$, the sample is classified into class 1, and class 2 otherwise. The decision boundary is as follows:

\begin{align}
  \|W_1\| \cos{\theta_1} = \|W_2\| \cos{\theta_2}
\label{eq:softmax-boundary}
\end{align}

\begin{figure}
\centering
\includegraphics[width=.8\linewidth]{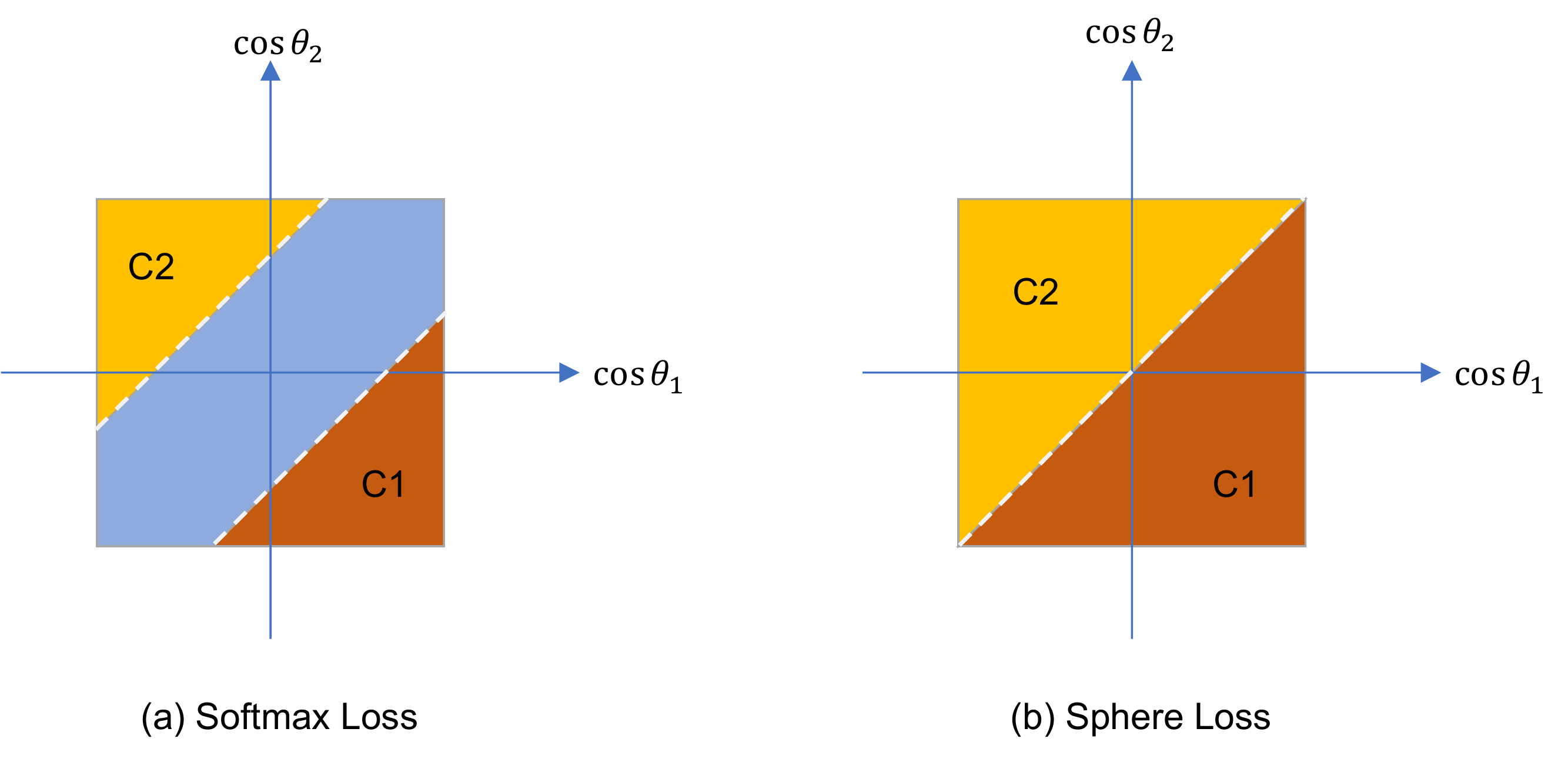}
\caption{The decision boundary of (a)Softmax Loss and (b)Sphere Loss. Samples lie within yellow area will be classified into class 2 and class 1 when samples lie within red area. Blue area means the class is uncertain because both angle and norm contribute to the decision.}
\label{fig:boundary}
\end{figure}

Equation \ref{eq:softmax-boundary} shows that both the norm and angle influence the final decision. As shown in Fig.~\ref{fig:boundary}, there is an intersection area of class 1 and class 2, and thus samples of two classes can not be distinguished only by the angle.

\subsection{Sphere Loss}
To eliminate the influence of norm and learn angularly discriminative features, we fix $\|W_j\| = 1$ and $\|x\|=1$ by L2 normalization as follows:

\begin{align}
  W_j  = \frac{W_j^*}{\|W_j^*\|}, x  = \frac{x^*}{\|x^*\|}
\end{align}

Where $W_j^*$ and $x^*$ are the original weight vector and feature vector.

In the original softmax loss function without normalization, when the angle between the feature vector and weight vector is the same, a sample tends to be classified into classes with larger norm, which we call weight bias, and a sample with larger norm tends to output a larger score, which we call feature bias. With the introduction of weight normalization and feature normalization, weight bias and feature bias are removed.

As shown in Fig.~\ref{fig:interpretation}(b), after normalization, the weight vector and feature vector are all mapped onto a hypersphere manifold, and the classification results only depend on the angle between the feature vector and weight vector.  For a binary classification, when $\cos{\theta_1}>\cos{\theta_2}$, the sample is classified into class 1, and class 2 otherwise. The decision boundary is:

\begin{align}
  \cos{\theta_1}=\cos{\theta_2}
\end{align}

As shown in Fig.~\ref{fig:boundary}(b), compared with softmax, there is a clear decision boundary and classification results only depend on the angle.

Combining weight normalization and feature normalization, we also add a scale factor to control the temperature of the softmax function. Note that $\|W_j\|=1$ and $\|x\|=1$ and we then get the Sphere Loss:

\begin{align}
  L_{sphere} = -\frac{1}{N}\sum_{i=1}^{N}\log\frac{e^{s\cos{\theta_{y_i}}}}{\sum_{j=1}^{C}e^{s\cos{\theta_j}}}
\label{eq:sphere}
\end{align}

where  $s$ is the scale factor. In this paper, we use $s = 14$ for all experiments. Equation~\ref{eq:sphere} is similar to the normalized
version of softmax loss (NSL) proposed in \cite{CosFace_2018}, but in \cite{CosFace_2018}, it is only an intermediate result proposed for the face recognition task and its effects are not fully exploited. Equation~\ref{eq:sphere} also matches the special case of additive margin softmax loss \cite{AAM_Softmax_2018} and additive angular margin loss \cite{ArcFace_2018} when the margin is set to 0.

With the supervision of Sphere Loss, we can learn an embedding on a hypersphere manifold, and different samples are discriminated by angles.

\subsection{Balanced Sampling Strategy}

A ReID datasets consists of images from different person where every person has an indefinite number of images. Usually there is no constraint on the proportion of different persons in a mini-batch and training data is chosen randomly from all the training images. However, the training process suffers from an imbalance of data. The network trains more on a person with more images, while it trains less on a person with less images. Thus the model tends to fit more on the person with more images. However, there is an a priori that every person is of the same importance and should be treated equally.

Therefore, we introduce a balanced sampling strategy. To generate a mini-batch, we randomly choose $P$ different persons without replacement, and for each person, we randomly choose $K$ images. Thus there are a total of $PK$ images in a mini-batch. For people with less than $K$ images, we use sampling with replacement, and sampling without replacement otherwise. After all persons are sampled, we say that an epoch is considered done.

This balanced sampling strategy is similar to the strategy proposed in \cite{TripletHard} for hard triplets mining, whereas in this paper, we use it to remove the imbalance of classes.

In this manner, for persons with more images we just ignore the over abundance of images, while for persons with less images we may use the same images multiple times. This approach guarantees that every person ID has the same number of instances.

\subsection{SphereReID Network}

\begin{figure}
\centering
\includegraphics[width=.8\linewidth]{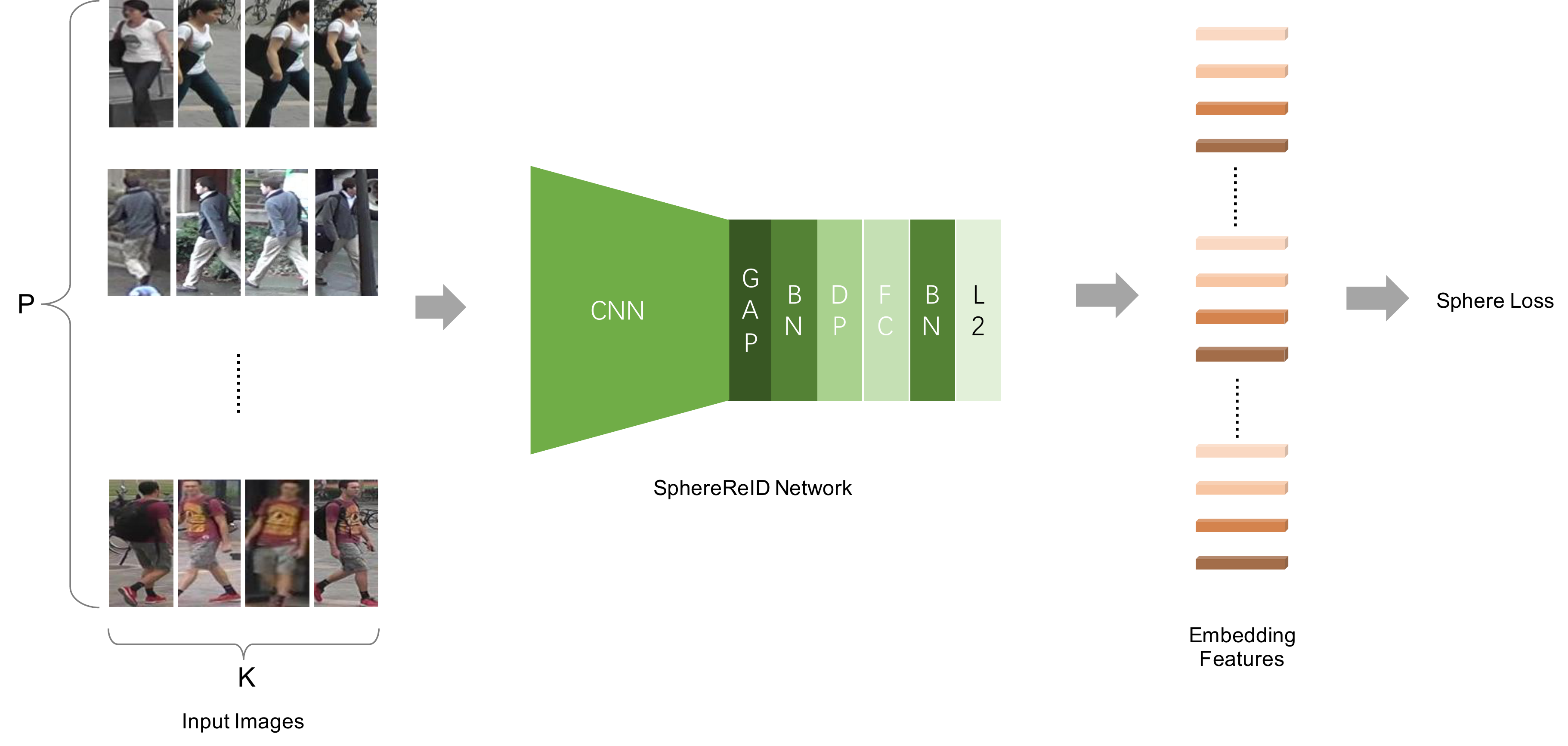}
\caption{The proposed SphereReID network structure. Inputs are a total of $PK$ images generated by a balanced sampling strategy. After the last convolutional layer of the ResNet-50 \cite{ResNet} backbone, a global average pooling (GAP), batch normalization (BN), dropout (DP), fully connected layer (FC), batch normalization (BN), L2 normalization (L2) are follows respectively.}
\label{fig:network}
\end{figure}

Combining Sphere Loss and the balanced sampling strategy, we propose a deep convolution neural network named SphereReID for the ReID task. As shown in Fig.~\ref{fig:network}, we use a ResNet-50 \cite{ResNet} network as the backbone network. After the last convolutional layer, a global average pooling follows to aggregate spatial information. Then we apply a batch normalization layer.

We also add a dropout layer as a regularizer, followed by a fully connected layer and another batch normalization layer. Now, we can apply weight normalization and feature normalization, and compute the Sphere Loss.

\section{Experiments}

\subsection{Datasets}

We conduct extensive experiments on four widely used ReID datasets: Market-1501 \cite{Market_1501_2015}, DukeMTMC-reID \cite{DukeMTMC_reID_2017}, CUHK03 \cite{CUHK_03}, and CUHK-SYSU \cite{CUHK_SYSU}.

\textbf{Market-1501} contains 32,668 annotated bounding box images of 1,501 labelled persons captured by five high-resolution cameras and one low-resolution camera. The dataset employes the Deformable Part Model (DPM) \cite{DPM} as the pedestrian detector. A total of 751 persons are used for training.

\textbf{DukeMTMC-reID} is a subset of Duke-MTMC \cite{Duke_MTMC} for person re-identification. It contains 36,411 annotated bounding box images of 1,812 different identities captured by eight high-resolution cameras. A total of 1,404 identities are observed by at least two cameras, and the remaining 408 identities are distractors. The training set contains 16,522 images of 702 identities and the test set contains the other 702 identities.

\textbf{CUHK03} contains 14,096 annotated bounding box images of 1,467 identities. Each identity is observed by two disjoint camera views. There are two kinds of bounding boxes available: one is manually cropped and the other is automatically detected by DPM \cite{DPM}. In this paper, we use the manually cropped version.

\textbf{CUHK-SYSU} containing 18,184 full images and 8,432 identities. A total of 99,809 bounding box images are annotated from full images. The training set contains 11,206 full images
and 5,532 persons, whereas the test set contains 6,978 full images of 2,900 persons.

\subsection{Implementation Details}
Our SphereReID model is built on the PyTorch framework. The backbone network is ResNet-50 \cite{ResNet} pre-trained on ImageNet and the original fully connected layer is discarded.

The inputs images are resized to $288\times144$ then randomly cropped to $256\times128$. The parameters $P$ and $K$ in the balanced sampling strategy are 16 and 4 respectively, as a result, a mini-batch size of 64 is used in our experiments.

We use the Adam optimizer with the default hyper-parameters($\epsilon=10^{-8}$, $\beta_1=0.9$, $\beta_2=0.99$). We set the initial learning rate to  $10^{-3}$ and apply the decay schedule at
epoch 80 and reduce the learning rate to $10^{-4}$. At epoch 100, we reduce the learning rate again to $10^{-5}$. The total number of training epochs for all conducted experiments is set to 140.

\begin{figure}
\centering
\includegraphics[width=.65\linewidth]{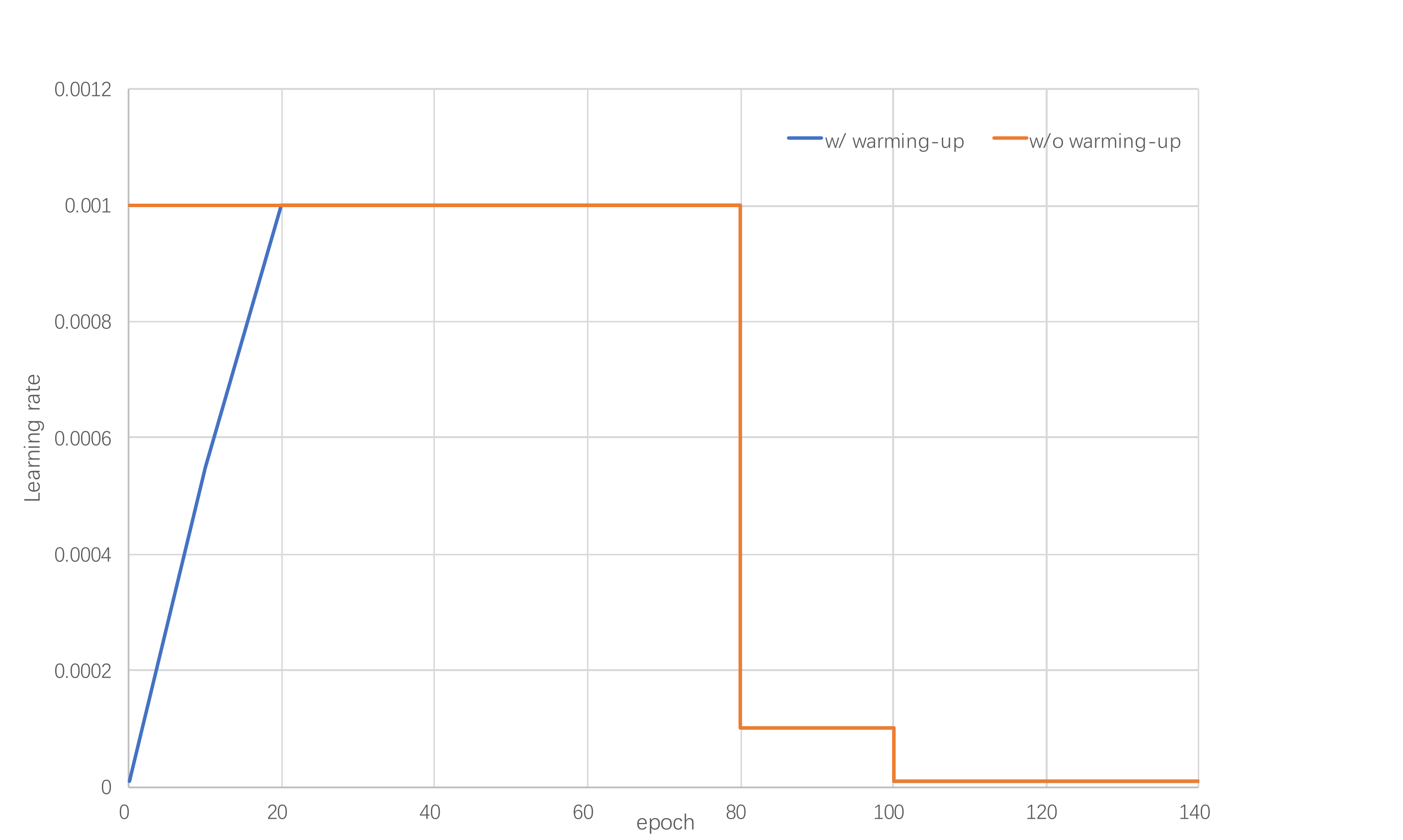}
\caption{Our learning rate schedules with or without warming-up. Blue one is with warming-up and the orange one is without warming-up.}
\label{fig:lr}
\end{figure}

We also introduce a warming-up strategy to bootstrap the network, as shown in Fig.~\ref{fig:lr}. We spend 20 epochs to linearly increase the learning rate from $5\times10^{-5}$ to $10^{-3}$. We think this strategy will help the network to initialize well before applying a large learning rate to optimize it. The experiment results are shown in the next section and demonstrate the effectiveness of this strategy.

\subsection{Results of SphereReID}
In this section, we go over different experiments settings and compare the rank-1 accuracies on Market-1501 \cite{Market_1501_2015}, DukeMTMC-reID \cite{DukeMTMC_reID_2017}, CUHK03 \cite{CUHK_03}, and CUHK-SYSU \cite{CUHK_SYSU}.

\subsubsection{Network Structure and Loss.}
After the last convolutional layer of the ResNet-50 \cite{ResNet} backbone, we have four different structures as follows: (A) global average pooling; (B) global average pooling, then a fully connected layer; (C) global average pooling, then a fully connected layer and a batch normalization; (D) global average pooling, batch normalization, dropout, fully connected layer and then a batch normalization again. The embedding feature size is 2048 for network-A and is 1024 for network-B, network-C and network-D. For network-D, the ratio of dropout is set to 0.5. Finally, L2 normalization is  applied for all the networks.

The results are shown in Table.~\ref{table:structure}. We can see that network-B is much better than network-A, which suggests that the additional fully connected layer can fuse input information and produce better embedding features. Network-C is also better than network-B, which demonstrates the effect of batch normalization. Network-D is the best and achieves 93.1\% rank-1 accuracy on Market-1501 which combines the strength of batch normalization and dropout. Table.~\ref{table:structure} also shows that sphere Loss is clearly better than softmax.

The subsequent experiments all use the network-D structure.

\begin{table}
\begin{center}
\caption{Results of different network structures.}
\label{table:structure}
\renewcommand{\arraystretch}{1.2}
\begin{tabular}{c|p{.1\textwidth}<{\centering}|p{.1\textwidth}<{\centering}|p{.1\textwidth}<{\centering}|p{.1\textwidth}<{\centering}|p{.1\textwidth}<{\centering}|p{.1\textwidth}<{\centering}|p{.1\textwidth}<{\centering}|p{.1\textwidth}<{\centering}}
\hline
\multirow{2}{*}{Network} & \multicolumn{2}{c|}{Market-1501} & \multicolumn{2}{c|}{DukeMTMC-reID} & \multicolumn{2}{c|}{CUHK03} & \multicolumn{2}{c}{CUHK-SYSU}\\
\cline{2-9}
&Sphere & Softmx & Sphere & Softmx & Sphere & Softmx & Sphere & Softmx \\

\hline
network-A & 59.5 & 57.9 & 49.1 & 46.1 & 67.3 & 62.3 & 77.4 & 78.1 \\
network-B & 86.4 & 65.6 & 75.5 & 53.9 & 86.8 & 63.7 & 90.6 & 83.1 \\
network-C & 92.3 & 72.7 & 81.6 & 59.0 & 91.5 & 62.2 & 94.1 & 83.1 \\
network-D & 93.1 & 77.3 & 81.9 & 61.9 & 92.8 & 66.6 & 94.3 & 86.2 \\
\hline
\end{tabular}
\end{center}
\end{table}

\subsubsection{Balanced Sampling Strategy.} The balanced sampling strategy can guarantee that each training identity has the same number of instances and alleviates the imbalance of sample size. As shown in Table.~\ref{table:balance-warmup}, when the model is trained with the balanced sampling strategy, the final performance is significantly boosted by a large margin, even with the exactly same network structure. This strategy may be applied to a wider area of tasks, helping to eliminate the class bias in a imbalanced dataset.

\begin{table}
\begin{center}
\caption{The influence of balanced sampling strategy and warming-up strategy.}
\label{table:balance-warmup}
\renewcommand{\arraystretch}{1.2}
\begin{tabular}{c|c|c|c|c}
\hline
  & Market-1501 & DukeMTMC-reID & CUHK03 & CUHK-SYSU\\
\hline
balanced, w/ warming-up  & 93.1 & 81.9 & 92.8 & 94.3 \\
imbalanced, w/ warming-up & 79.3 & 56.5 & 79.2 & 89.9 \\
balanced, w/o warming-up & 77.1 & 64.4 & 80.6 & 88.9 \\
\hline
\end{tabular}
\end{center}
\end{table}

\subsubsection{Influence of Warming-up.} As shown in Table.~\ref{table:balance-warmup}, with a warming-up process of the learning rate to help the network bootstrap, rather than applying a large learning rate from the beginning,  the network can converge on a much better point.
It is intuitive that when a network is initialized by weights pre-trained on ImageNet and never be used for the ReID task, a large learning rate may be inappropriate. The proposed fine-to-coarse then coarse-to-fine learning rate schedule can help set up a better initialization status and thus result in a better performance. The proposed warming-up strategy is not limited to the ReID task, and it may be applied to other areas to obtain a better optimizing result.

\subsubsection{Ratio of Dropout.} We try three different ratios of dropout, and a network without dropout. Results are shown in Table.~\ref{table:dropout}. We can see that the networks with no dropout(when the ratio is 0) or with too much dropout are inferior to the network with a modest dropout ratio of 0.25. In the following experiments, we will fix the dropout ratio to 0.25.

\begin{table}
\begin{center}
\caption{The influence of dropout ratio.}
\label{table:dropout}
\renewcommand{\arraystretch}{1.2}
\begin{tabular}{c|c|c|c|c}
\hline
ratio & Market-1501 & DukeMTMC-reID & CUHK03 & CUHK-SYSU\\
\hline
0 & 93.1 & 82.2 & 92.5 & 94.6 \\
0.25 & 92.8 & 83.5 & 93.2 & 94.8 \\
0.5 & 93.1 & 81.9 & 92.8 & 94.3 \\
0.75 & 91.3 & 80.5 & 1.5 & 93.5 \\

\hline
\end{tabular}
\end{center}
\end{table}

\subsubsection{Influence of the Bias Term.} In the last fully connected layer, the bias term $b$ can be set to 0 or learned automatically. We train two networks with and without the automatically learned bias term. Results are shown in Table.~\ref{table:bias}. We can see that the network with the bias term automatically learned performs slightly better than the network without the bias term.

From now on, we will refer this best network setting with a bias term as SphereReID network.

\begin{table}
\begin{center}
\caption{The influence of bias term in the last fully connected layer.}
\label{table:bias}
\renewcommand{\arraystretch}{1.2}
\begin{tabular}{c|c|c|c|c}
\hline
  & Market-1501 & DukeMTMC-reID & CUHK03 & CUHK-SYSU\\
\hline
w/ bias & 93.7 & 83.9 & 92.6 & 94.9 \\
w/o bias & 92.8 & 83.5 & 93.2 & 94.8 \\
\hline
\end{tabular}
\end{center}
\end{table}

\subsubsection{Test Image Size.} In the training phase, we resize the image to $288\times144$, then randomly crop it to $256\times128$. In the testing phase, results of different image sizes are shown in Table.~\ref{table:size}. We can see that with larger input size, the performance is better on Market-1501, CUHK03, and CUHK-SYSU, and is worse on DukeMTMC-reID. After examining images from the four datasets, we find that images from DukeMTMC-reID have a larger border background area. Thus, we use $256\times128$ test size for DukeMTMC-reID and $288\times144$ for the others.

\begin{table}
\begin{center}
\caption{The influence of test image size. Images are resized to $288\times144$ with and without center crop of $256\times128$.}
\label{table:size}
\renewcommand{\arraystretch}{1.2}
\begin{tabular}{c|c|c|c|c}
\hline
  & Market-1501 & DukeMTMC-reID & CUHK03 & CUHK-SYSU\\
\hline
$256\times128$ & 93.7 & \textbf{83.9} & 92.6 & 94.9 \\
$288\times144$ & \textbf{94.4} & 82.7 & \textbf{93.1} & \textbf{95.4} \\
\hline
\end{tabular}
\end{center}
\end{table}

\subsection{Comparison with the State of the Art}

We compare SphereReID with the state of the art. As shown in Table.~\ref{table:market}, Table.~\ref{table:cuhk-sysu}, and Table.~\ref{table:cuhk03}, our single mode consistently outperforms the state of the art in terms of both accuracy and mAP, and it achieves 94.4\% rank-1 accuracy on Market-1501. It is necessary to point out that no extra attributes, skeleton datasets, or models are used in our SphereReID network.

On DukeMTMC-reID, as shown in Table.~\ref{table:duke}, PCB+RPP \cite{Refined_2017} obtains competitive results, but it is trained by a three stage process with fine-tuning. However, the proposed SphereReID is trained end-to-end without fine-tuning and is clearly better than PCB+RPP on Market-1501. Furthermore, SphereReID achieves all the results with a feature size of 1,024, while PCB+RPP uses a feature size of 12,288, which proves that our SphereReID features mapped onto a hypersphere manifold are more discriminative.

\begin{table}
\begin{center}
\caption{Comparison with the State of the Art on Market-1501}
\label{table:market}
\renewcommand{\arraystretch}{1.2}
\begin{tabular}{c|cccc}
\hline
method & rank-1 & rank-5 & rank-10 & mAP \\
\hline
Spindle \cite{Spindle_2017} & 76.9 & 91.5 & 94.6 & - \\
SVDNet \cite{SVDNet} & 82.3 & 92.3 & 95.2 & 62.1 \\
PDC \cite{PoseDriven_2017} & 84.1 & 92.7 & 94.9 & 63.4 \\
Mutual \cite{Mutual} & 87.7 & - & -& 68.8 \\
PSE \cite{Re_Ranking_2} & 87.7 & 94.5 & 96.8 & 69.0 \\
PartLoss \cite{PartLoss_2017} & 88.2 & - & - & 69.3 \\
DPFL \cite{DPFL} & 88.9 & - & - & 72.6 \\
CamStyle \cite{CamStyle} & 89.5 & - & - & 71.6 \\
GLAD \cite{GLAD_2017} & 89.9 & - & - & 73.9 \\
HA-CNN \cite{HA-CNN} & 91.2 & - & - & 75.7 \\
Deep-Person \cite{DeepPerson_2017} & 92.3 & - & - & 79.6 \\
PCB+RPP \cite{Refined_2017} & 93.8 & 97.5 & 98.5 & 81.6 \\
\hline
SphereReID & \textbf{94.4} & \textbf{98.0} & \textbf{98.7} & \textbf{83.6} \\
\hline
\end{tabular}
\end{center}
\end{table}

\begin{table}
\begin{center}
\caption{Comparison with the State of the Art on CUHK-SYSU}
\label{table:cuhk-sysu}
\renewcommand{\arraystretch}{1.2}
\begin{tabular}{c|cccc}
\hline
method & rank-1 & rank-5 & rank-10 & mAP \\
\hline
deep \cite{CUHK_SYSU} & 62.7 & - & - & 55.7 \\
DLDP \cite{DLDP} & 76.7 & - & - & 74.0 \\
NPSM \cite{NPSM} & 81.2 & - & - & 77.9 \\
\hline
SphereReID & \textbf{95.4} & \textbf{98.6} & \textbf{98.9} & \textbf{93.9} \\
\hline
\end{tabular}
\end{center}
\end{table}

\begin{table}[h]
\begin{center}
\caption{Comparison with the State of the Art on DukeMTMC-reID}
\label{table:duke}
\renewcommand{\arraystretch}{1.2}
\begin{tabular}{c|cccc}
\hline
method & rank-1 & rank-5 & rank-10 & mAP \\
\hline
SVDNet \cite{SVDNet} & 76.7 & 86.4 & 89.9 & 56.8 \\
HA-CNN \cite{HA-CNN} & 78.3 & - & - & 57.6 \\
DPFL \cite{DPFL} & 79.2 & - & - & 60.6 \\
PSE \cite{Re_Ranking_2} & 79.8 & 89.7 & 92.2 & 62.0 \\
HA-CNN \cite{HA-CNN} & 80.5 & - & - & 63.8 \\
Deep-Person \cite{DeepPerson_2017} & 80.9 & - & - & 64.8 \\
PCB+RPP \cite{Refined_2017} & 83.3 & 90.5 & \textbf{92.5} & \textbf{69.2} \\
\hline
SphereReID & \textbf{83.9} & \textbf{90.6} & 92.4 & 68.5 \\
\hline
\end{tabular}
\end{center}
\end{table}

\begin{table}[h!]
\begin{center}
\caption{Comparison with the State of the Art on CUHK03}
\label{table:cuhk03}
\renewcommand{\arraystretch}{1.2}
\begin{tabular}{c|ccc}
\hline
method & rank-1 & rank-5 & rank-10 \\
\hline
PartLoss \cite{PartLoss_2017} & 82.8 & 96.6 & 98.6\\
GLAD \cite{GLAD_2017} & 85.0 & 97.9 & 99.1 \\
DPFL \cite{DPFL} & 86.7 & - & - \\
Spindle \cite{Spindle_2017} & 88.5 & 97.8 & 98.6 \\
PDC \cite{PoseDriven_2017} & 88.7 & 98.6 & 99.2 \\
Deep-Person \cite{DeepPerson_2017} & 91.5 & \textbf{99.0} & \textbf{99.5} \\
\hline
SphereReID & \textbf{93.1} & 98.7 & 99.4 \\
\hline
\end{tabular}
\end{center}
\end{table}

\section{Conclusions}

In this paper, we introduce a modified softmax loss function called Sphere Loss with weight normalization and feature normalization. We also propose a CNN network adopting Sphere Loss called SphereReID which can learn the feature embedding on a hypersphere manifold. We train the SphereReID end-to-end with the balanced sampling strategy and warming-up strategy and our single model outperforms the state of the art on all four datasets without re-ranking or fine-tuning.

To the best of our knowledge, this is the first network to learn a deep hypersphere manifold embedding for person re-identification, and the proposed SphereReID network demonstrates the effectiveness of this concept. We have provided a new idea for ReID and there are more further improvements can be explored by the person re-identification community, for example, the addition of a margin term to increase inter-class variation and reduce intra-class variation.

And the proposed warming-up strategy can further boost the performance of deep neural network without extra computing overhead. It's very simple to implement and can be easily introduced into the training process. In this paper, we focus on SphereReID for person re-identification task, but it can also be used in other tasks, which remains for the computer vison community to explorer in the further.

\section{Acknowledgement}
Our work was supported by the Public Projects of Zhejiang Province, China (No. LGF18F030002) and the National Natural Science Foundation of China (No. 61633019).

\bibliographystyle{splncs}
\bibliography{egbib}
\end{document}